\documentclass[runningheads]{llncs}

\usepackage{eccv}

\usepackage{eccvabbrv}

\usepackage{graphicx}
\usepackage{booktabs}

\usepackage[accsupp]{axessibility}  

\usepackage{graphicx}
\usepackage{booktabs}
\usepackage{amssymb}
\usepackage{pifont}
\usepackage{graphicx}
\newcommand{\cmark}{\ding{51}}%
\newcommand{\xmark}{\ding{55}}%
\usepackage{url}            
\usepackage{booktabs}       
\usepackage{amsfonts}       
\usepackage{nicefrac}       
\usepackage{microtype}      
\usepackage{graphicx}
\usepackage{float}
\usepackage{tabularx, makecell, multirow} 

\usepackage{hyperref}
\usepackage{orcidlink}

\begin{document}

\title{PACE: A Large-Scale Dataset with Pose Annotations in Cluttered Environments}

\titlerunning{PACE: Pose Annotations in Cluttered Environments}

\author{Yang You\orcidlink{0000-0003-0125-0792}\textsuperscript{1}$^\star$ \hspace{.1cm} Kai Xiong\textsuperscript{2} \hspace{.1cm} Zhening Yang\textsuperscript{3} \hspace{.1cm} Zhengxiang Huang\textsuperscript{2} \\Junwei Zhou\textsuperscript{2} \hspace{.1cm}  Ruoxi Shi\textsuperscript{4} \hspace{.1cm} Zhou Fang\textsuperscript{2}  \\Adam W. Harley\orcidlink{0000-0002-9851-4645}\textsuperscript{1} \hspace{.1cm} Leonidas Guibas\orcidlink{0000-0002-8315-4886}\textsuperscript{1} \hspace{.1cm} Cewu Lu\textsuperscript{2}\thanks{Corresponding authors. 
Email: \email{yangyou@stanford.edu}, \email{lucewu@sjtu.edu.cn}
} 
}

\authorrunning{Y.~You et al.}

\institute{Stanford University \and
Shanghai Jiao Tong University\\
\and
Horizon Robotics Inc.\\
\and
UC San Diego
}

\maketitle

\begin{figure*}[!ht]
    \centering
    \includegraphics[width=1.0\linewidth]{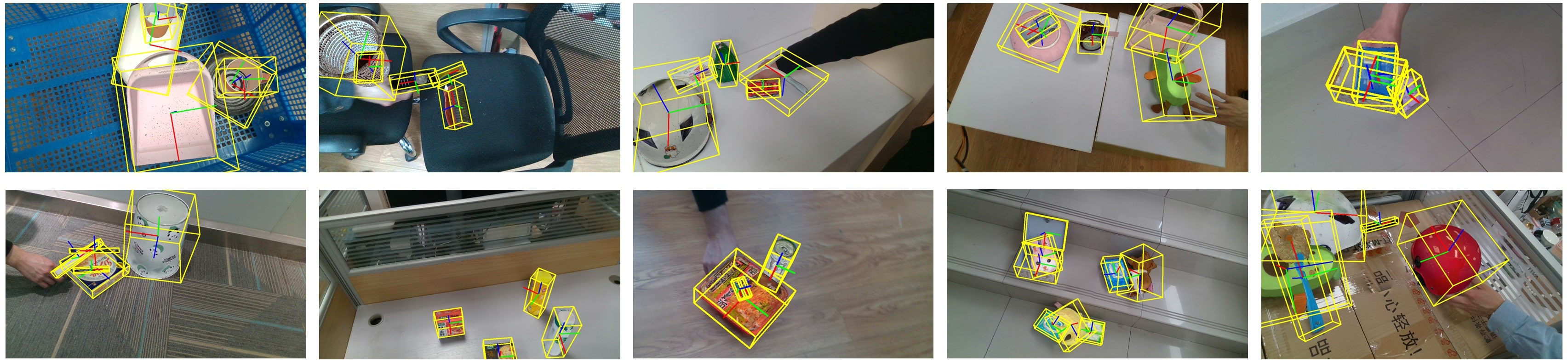}
    \caption{We propose PACE: a large-scale object pose dataset, with diverse objects, complex scenes, and various types of occlusions, reflecting real-world challenges. 
    }
    \label{fig:intro}
\end{figure*}

\begin{abstract}
We introduce \textbf{PACE} (Pose Annotations in Cluttered Environments), a large-scale benchmark designed to advance the development and evaluation of pose estimation methods in 
cluttered scenarios. PACE provides a large-scale real-world benchmark for both instance-level and category-level settings. The benchmark consists of 55K frames with 258K annotations across 300 videos, covering 238 objects from 43 categories and featuring a mix of rigid and articulated items in cluttered scenes. To annotate the real-world data efficiently, we develop an innovative annotation system with a calibrated 3-camera setup. Additionally, we offer \textbf{PACE-Sim}, which contains 100K photo-realistic simulated frames with 2.4M annotations across 931 objects.
We test state-of-the-art algorithms in PACE along two tracks: pose estimation, and object pose tracking, revealing the benchmark's challenges and research opportunities. 
Our benchmark code and data is available on \url{https://github.com/qq456cvb/PACE}.
\end{abstract}

\section{Introduction}
\label{sec:intro}
The field of 3D object pose estimation is integral to a myriad of applications, particularly within robotic manipulation. Recent advancements in both instance and category-level pose estimation have been bolstered by deep learning approaches, and perhaps more importantly, by \textit{data}. 

PoseCNN~\cite{xiang2017posecnn} advanced pose estimation into the deep learning era, and simultaneously introduced the influential YCB-Video dataset. This benchmark has catalyzed methodological development and offered a consistent evaluation platform. Besides, the Benchmark for 6D Object Pose Estimation (BOP) challenges~\cite{hodavn2020bop} have consolidated datasets metrics for instance-level pose estimation.

\begin{figure}[!ht]

    \centering
    \begin{minipage}{0.5\linewidth}
        \centering 
        \includegraphics[width=\linewidth]{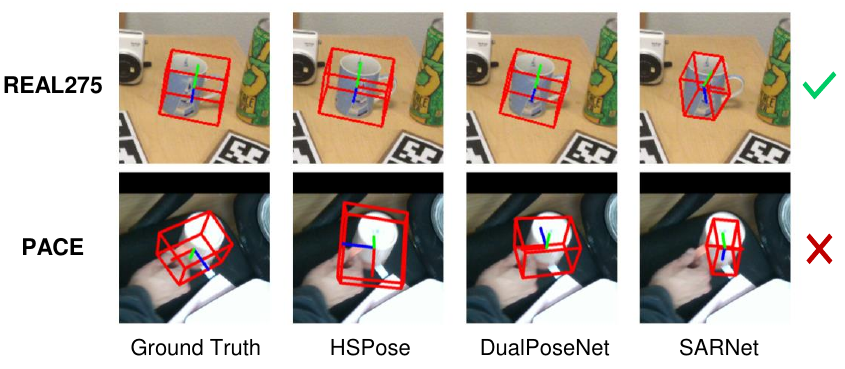} 
    \end{minipage}
    \hfill 
    \begin{minipage}{0.45\linewidth}
        \centering 
        \resizebox{\linewidth}{!}{%
        \begin{tabular}{llcccc}
            \toprule
            \multirow{2}*{Dataset} & \multirow{2}*{Method} & \multicolumn{4}{c}{AP@15$^\circ$5cm (\%)$\uparrow$} \\
            \cmidrule(lr){3-6}
            ~ & ~ & bottle & bowl & can & mug \\
            \midrule
            \multirow{3}*{REAL275~\cite{wang2019normalized}} & HS-Pose~\cite{zheng2023hs} & 99.8 & 99.8 & 99.5 & 86.2 \\
            ~ & DualPoseNet~\cite{lin2021dualposenet} & 97.5 & 99.7 & 97.7 & 68.7 \\
            ~ & SARNet~\cite{lin2022sar} & 98.2 & 98.2 & 97.4 & 70.6 \\
            \midrule
            \multirow{3}*{PACE (Ours)} & HS-Pose~\cite{zheng2023hs} & 0.4 & 0.0 & 0.8 & 7.5 \\
            ~ & DualPoseNet~\cite{lin2021dualposenet} & 0.3 & 2.0 & 0.5 & 1.6 \\
            ~ & SARNet~\cite{lin2022sar} & 2.1 & 0.0 & 0.3 & 0.0 \\
            \bottomrule
        \end{tabular}
        }
    \end{minipage}
    \caption{While current state-of-the-art methods yield satisfactory outcomes on the NOCS-REAL275 dataset, their models' performance significantly deteriorates when transferred to previously unseen datasets such as PACE. \textbf{Left:} Qualitative visualizations of various models' pose predictions for a mug in REAL275 vs. a mug in PACE. \textbf{Right:} The performance of state-of-the-art methods markedly declines on PACE, even when evaluating on categories that exist in both datasets.} 
    \label{figure:nocs}
    
\end{figure}

In parallel, NOCS~\cite{wang2019normalized} has addressed category-level pose estimation, albeit with a smaller real-world dataset for validation and testing. Despite these strides, the field grapples with a fundamental challenge: state-of-the-art pose estimation models still perform poorly in real-world settings, and existing evaluation datasets are too constrained to thoroughly reveal this fact. 
The NOCS-REAL275 dataset, for instance, spans only six categories and includes a mere 18 videos. This limitation has led to an array of published models which perform near-perfectly on NOCS but do not generalize to new data, as demonstrated in Figure \ref{figure:nocs}. 


In this work, we introduce PACE, a benchmark for pose estimation, and present a comprehensive study evaluating a wide range of pose estimation and tracking methods. Our contributions are threefold:

\begin{itemize}
    \item \textbf{The PACE dataset}: This dataset 
    includes 238 objects across 43 categories, captured in 300 video clips within diverse scenes. With an average of 183 frames per clip, the dataset consists of 55K frames and 258K annotations, 
    providing a large-scale benchmark 
    for pose estimation. We also provide PACE-Sim, a photo-realistic training dataset which contains 100K frames with 2.4M annotations across 931 objects.

    \item \textbf{Our evaluation study}: 
    To the best of our knowledge, we are the first to analyze and report the performance of state-of-the-art (SOTA) pose estimation methods in a large-scale cluttered setting. These results provide valuable insights on the scalability and generalizability of SOTA methods, making it clear that they are far from reliable in general. 

    \item \textbf{Our annotation pipeline}: We will open-source our annotation pipeline, which uses a calibrated 3-camera system, enhancing the precision and scalability of annotating poses in real data. This tool significantly mitigates human error and reduces the effort required for annotating 3D poses, providing a solution to one of the major bottlenecks in pose dataset creation.    
    
\end{itemize}

Overall, our work aims to support the development of more robust and generalizable pose estimation techniques, thereby facilitating progress towards successful pose estimation in the real world. 

\section{Related Works}
\label{sec:related}

The field of 3D object pose estimation has seen substantial progress over the past few years. This progress has been facilitated by the introduction of standardized datasets and the development of innovative algorithms.

\subsection{Object Pose Datasets}

\subsubsection{Instance-Level Pose Datasets}
YCB-Video dataset~\cite{xiang2017posecnn} is a comprehensive resource for 6D object pose estimation, containing a large number of video frames with accurate pose annotations for 21 objects.
LINEMOD-Occluded~\cite{brachmann2014learning} offers a challenging setting for pose estimation with piled multiple objects in occluded scenes.
NAVI dataset~\cite{jampani2023navi} presents casually captured images of objects with high-quality 3D scans and precise 2D-3D alignments for 3D reconstruction tasks. Recently, StereoObj-1M~\cite{liu2021stereobj} is proposed to address challenging cases such as object transparency, in addition to common challenges of occlusion and symmetry.

\subsubsection{Category-Level Pose Datasets}
NOCS-REAL275~\cite{wang2019normalized} and Wild6D~\cite{ze2022category} both annotate a limited number of categories with upright poses. Objectron~\cite{ahmadyan2021objectron} is a collection of short, single-object-centric video clips with no clutters. 
The Scan2CAD~\cite{avetisyan2019scan2cad} dataset aligns 14225 CAD models from ShapeNet to 1506 ScanNet scans, promoting CAD model alignment in RGB-D scans. Pix3D~\cite{sun2018pix3d} is a benchmark with image-shape pairs and pixel-level 2D-3D alignment, aiding in shape reconstruction and retrieval. 
HANDAL~\cite{guo2023handal} focuses on pose estimation and affordance prediction for robotics-ready manipulable objects. 
HouseCat6D~\cite{jung2024housecat6d} encompasses 194 diverse objects across 10 household categories with additional grasp annotations.
ROPE~\cite{zhang2024omni6dpose} provides a large-scale evaluation benchmark covering a wide spectrum of categories. However, it does not include challenging scenarios such as moving and piled objects. Table~\ref{tab:comp} presents a comparative comparison of these datasets, alongside our new dataset PACE.

\begin{table*}[ht]
    \centering
    \resizebox{\linewidth}{!}{
    \begin{tabular}{lcccccccccccc}
    \toprule
         &  Modality & Cat. & Obj. & Vid. & Img. & Anno. & CAD & Mov. & Occ. & Marker-free & Artic. & Piled\\
         \midrule
         YCB-Video~\cite{xiang2017posecnn} & RGBD & - & 21 & 12 & 20K & 99K & \textcolor{green}{\cmark} & \textcolor{red}{\xmark} & \textcolor{green}{\cmark} & \textcolor{green}{\cmark} & \textcolor{red}{\xmark} & \textcolor{green}{\cmark}\\
       LINEMOD-O~\cite{brachmann2014learning} & RGBD & - & 8 & 1 & 1.2K & 9.2K & \textcolor{green}{\cmark} & \textcolor{red}{\xmark} & \textcolor{green}{\cmark} & \textcolor{red}{\xmark} & \textcolor{red}{\xmark} & \textcolor{green}{\cmark}\\
       NAVI~\cite{jampani2023navi} & RGBD & - & 36 & 324 & 10K & 10k & \textcolor{green}{\cmark} & \textcolor{red}{\xmark} & \textcolor{red}{\xmark} & \textcolor{green}{\cmark} & \textcolor{red}{\xmark} & \textcolor{red}{\xmark}\\
       StereoObj-1M~\cite{liu2021stereobj} & RGBD & - & 18 & 182 & 393K & 1.5M & \textcolor{green}{\cmark} & \textcolor{red}{\xmark} & \textcolor{green}{\cmark} & \textcolor{red}{\xmark} & \textcolor{red}{\xmark} & \textcolor{green}{\cmark}\\
       \midrule
       NOCS-REAL275~\cite{wang2019normalized}  & RGBD & 6 & 42 & 18 & 8K &  & \textcolor{green}{\cmark} & \textcolor{red}{\xmark} & \textcolor{green}{\cmark} & \textcolor{red}{\xmark} & \textcolor{red}{\xmark} & \textcolor{red}{\xmark}\\
       Wild6D~\cite{ze2022category} & RGBD & 5 & 1722 & 5166 & 1.1M & 1.1M & \textcolor{red}{\xmark} & \textcolor{red}{\xmark} & \textcolor{red}{\xmark} & \textcolor{green}{\cmark} & \textcolor{red}{\xmark} & \textcolor{red}{\xmark}\\
       Objectron~\cite{ahmadyan2021objectron} & RGB & 9 & 17k & 14k & 4M & 4M & \textcolor{red}{\xmark} & \textcolor{red}{\xmark} & \textcolor{red}{\xmark} & \textcolor{green}{\cmark} & \textcolor{red}{\xmark} & \textcolor{red}{\xmark}\\
       Scan2CAD~\cite{avetisyan2019scan2cad} & RGBD & 9 & 3K & 1506 & - & 14K & \textcolor{green}{\cmark} & \textcolor{red}{\xmark} & \textcolor{green}{\cmark} & \textcolor{green}{\cmark} & \textcolor{red}{\xmark} & \textcolor{red}{\xmark}\\
       Pix3D~\cite{sun2018pix3d} & RGBD & 9 & 395 & - & 10K & 10K & \textcolor{green}{\cmark} & \textcolor{red}{\xmark} & \textcolor{red}{\xmark} & \textcolor{green}{\cmark} & \textcolor{red}{\xmark} & \textcolor{red}{\xmark}\\
       HANDAL~\cite{guo2023handal} & RGB & 17 & 212 & 2K & 308K & 308K & \textcolor{green}{\cmark} & \textcolor{green}{\cmark} & \textcolor{green}{\cmark} & \textcolor{green}{\cmark} & \textcolor{red}{\xmark} & \textcolor{red}{\xmark}\\
       HouseCat6D~\cite{jung2024housecat6d} & RGBD & 10 & 192 & 41 & 24K & 160K & \textcolor{green}{\cmark} & \textcolor{red}{\xmark} & \textcolor{green}{\cmark} & \textcolor{green}{\cmark} & \textcolor{red}{\xmark} & \textcolor{green}{\cmark}\\
       ROPE~\cite{zhang2024omni6dpose} & RGBD & 149 & 581 & 363 & 332K & 1.5M & \textcolor{green}{\cmark} & \textcolor{red}{\xmark} & \textcolor{green}{\cmark} & \textcolor{green}{\cmark} & \textcolor{red}{\xmark} & \textcolor{red}{\xmark}\\
       PACE (Ours) & RGBD & 43 & 238 & 300 & 55K & 258K & \textcolor{green}{\cmark} & \textcolor{green}{\cmark} & \textcolor{green}{\cmark} & \textcolor{green}{\cmark} & \textcolor{green}{\cmark} & \textcolor{green}{\cmark}\\
    \bottomrule
    \end{tabular}}
    
    \caption{\textbf{Comparison of object pose datasets.} From left to right, the table captures input modality, number of categories, number of instances, number of videos, number of images, number of total annotations, whether 3D CAD models are provided, whether videos include still and/or moving objects, whether objects are occluded in some frames, whether images contain artificial markers, whether poses for each part of articulated objects are provided, and whether multiple objects are piled in some frames. Compared with most other datasets, PACE contains moving and articulated objects. }
    \label{tab:comp}
    
\end{table*}

\subsection{Pose Estimation Methods}

\subsubsection{Instance-level Pose Estimation}
PPF (Point Pair Features)~\cite{drost2010model} is a pre-deep learning standard for instance-level pose estimation, using local geometric features from point clouds. 
PoseCNN~\cite{xiang2017posecnn} is the first model trained end-to-end for the task. 
CosyPose~\cite{labbe2020cosypose} integrates a global refinement strategy in its end-to-end pipeline. SurfEmb~\cite{haugaard2022surfemb} leverages surface embeddings for correspondence matching, and GDRNPP~\cite{Wang_2021_GDRN} uses geometry-guided regression. 

\subsubsection{Category-level Pose Estimation}
Category-level pose estimation extends the challenge to generic object categories. NOCS~\cite{wang2019normalized} introduces a unified coordinate space for all objects, predicting object NOCS maps from RGB images. SGPA~\cite{chen2021sgpa} aims to adapt the structure-guided prior in the pose estimation process, while SAR-Net~\cite{lin2022sar} uses shape alignment and symmetric correspondence to estimate a coarse 3D object shape and facilitate object center and size estimation. Recently, HS-Pose~\cite{zheng2023hs} proposes a network structure with a HS-layer that extends 3D graph convolution to extract hybrid scope latent features from point clouds for category-level object pose estimation. GenPose~\cite{zhang2023genpose} incorporates probabilistic stable diffusion models into pose estimation, achieving strong performance.

\subsection{Pose Tracking Methods}
\subsubsection{Instance-level Pose Tracking}
Methods like RBOT \cite{tjaden2018region} use RGB data and 3D models to track multiple objects, employing color histograms in their cost function. PoseRBPF \cite{deng2021poserbpf} separates rotation and translation, using an autoencoder for rotation feature embeddings. ICG \cite{stoiber2022iterative} iteratively refines pose using geometric cues and is effective for textureless objects, with extensions incorporating visual data \cite{stoiber2023fusing}. The first deep learning tracker, D6DT \cite{garon2017deep}, and se(3)-TrackNet \cite{wen2020se} predict frame-to-frame relative poses, using a render-and-compare strategy.

\subsubsection{Category-level Pose Tracking}
6-PACK \cite{wang20206} performs category-level tracking, using DenseFusion \cite{wang2019densefusion} features and an attention mechanism for unsupervised keypoint ordering and interframe motion via keypoint matching. 
BundleTrack \cite{wen2021bundletrack} performs pose tracking without relying on 3D models, instead using video segmentation and pose graph optimization. CAPTRA \cite{weng2021captra} tracks 9DoF poses for rigid and articulated objects, with heads for rotation regression and normalized coordinate prediction, from which 3D size and translation can be computed.

\section{Construction of PACE}

\begin{figure*}[ht]
    \centering
    \includegraphics[width=0.9\textwidth]{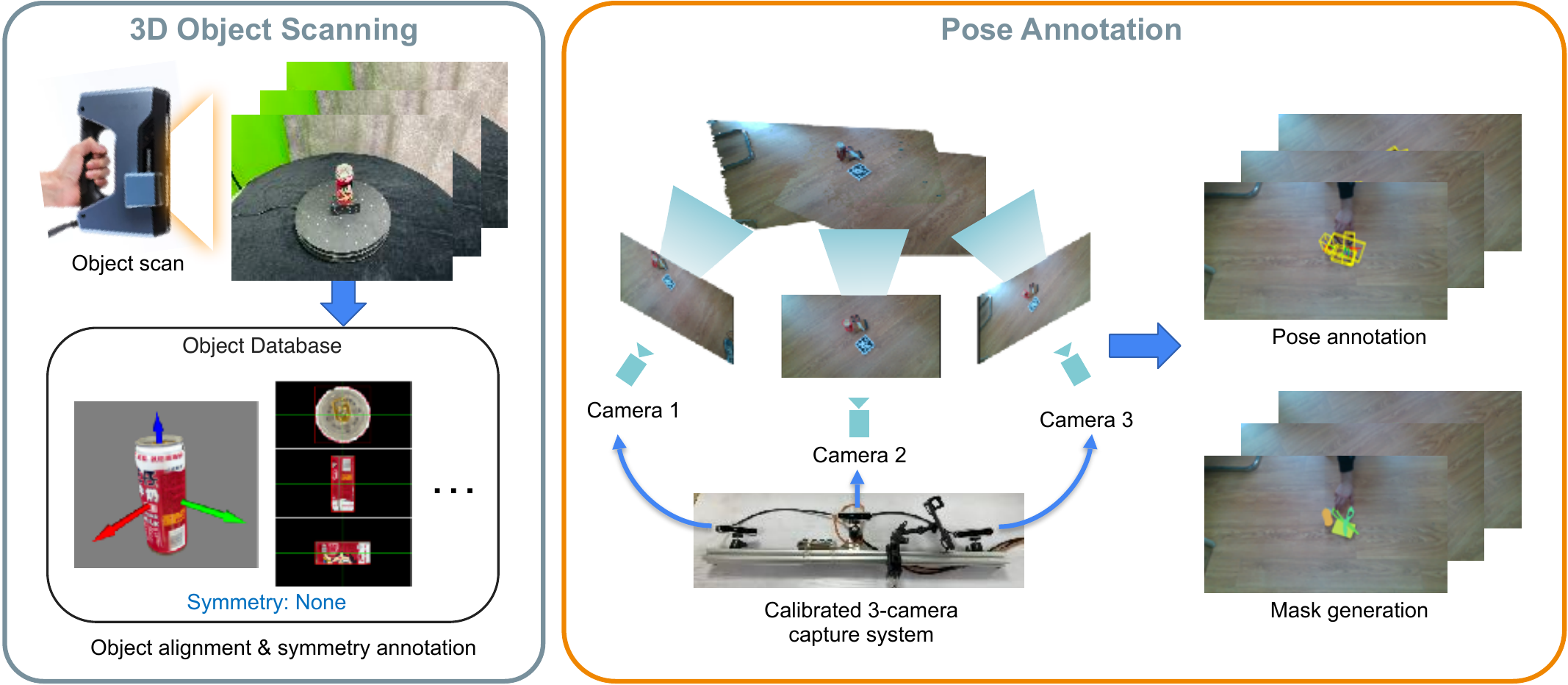}
    \caption{Overview of the PACE annotation pipeline.}
    \label{fig:pipeline}
    
\end{figure*}

A key contribution of this work is the establishment of a \textbf{scalable} and \textbf{reliable} annotation framework, enabling the collection of large-scale and accurate pose annotations. 
An overview of the pipeline is depicted in Figure~\ref{fig:pipeline}. 

\subsection{Acquisition of 3D Scans of Common Objects}
We began by digitizing an extensive collection of commonplace objects. These items were categorized into 43 distinct categories, as represented in Figure~\ref{fig:objdist}.
The Einscan Pro 2X was utilized for rapid scanning of all objects, typically completing within 5 to 10 minutes per object. To expedite this process, we employed a rotatable platform to acquire multiple viewpoints of objects. 
After scanning, objects were manually aligned to a standard pose within a uniform coordinate system. For each object, we centered its axis-aligned bounding box at the origin and then aligned the bounding box orientation along a common axis within the category. We additionally annotated rotational symmetries. 
High-resolution meshes were simplified to lower-resolution ones for a smoother annotation workflow.

%

\paragraph{Articulated Objects}
Diverging from many prior pose estimation datasets, our collection encompasses a very wide set of objects, including \textbf{articulated objects} from the AKB48~\cite{liu2022akb} dataset, namely: {scissors}, {cutters}, {clips}, and {boxes}. We adopt the alignment methodology from the original AKB48 dataset, without modification. These objects are segmented into multiple parts with hierarchical relationships, presenting a difficult challenge for pose estimation.

\subsection{RGB-D Sequence Acquisition}
We implemented a 3-camera system to aid in data acquisition and annotation, comprising three Intel Realsense D415 RGB-D cameras affixed to a metal framework, as illustrated in the bottom of Figure~\ref{fig:pipeline}. The advantages of this setup include: 
  (1) tripling the data yield; 
   (2) reducing ambiguity in pose annotation, especially regarding translation along the depth dimension, by enforcing annotation consistency across all 3 views;
   (3) enhancing tracking accuracy of still objects with ArUco markers~\cite{garrido2014automatic}, making PnP more stable. 

\paragraph{Calibration of Multi-Camera Extrinsic Parameters}
We calibrated this multi-camera system through a semi-automatic process. 
ArUco markers~\cite{garrido2014automatic} 
alone proved insufficient for high-accuracy rotation estimation. Hence, we resorted to 
trifocal tensor estimation, i.e. TFT~\cite{julia2018critical}. The process begins with feature extraction and matching, followed by bundle adjustment to refine the positions of the 3D landmarks and camera poses. For reliable feature matching, we employed the SuperPoint~\cite{detone2018superpoint} descriptor and SuperGlue~\cite{sarlin2020superglue} matcher, using a stringent threshold for matching. We observed that rotational component of the resulting extrinsic parameters is precise, but the translation aspect suffers from scale ambiguity inherent in Structure-from-Motion approaches. We corrected this by calibrating the scale against markers, applying the following formula to obtain the optimal scale factor, by setting gradient of $\|s\hat{\mathbf{t}}-\mathbf{t}'\|_2^2$ to zero:

\[s_{1\rightarrow 2} = \frac{\mathbf{\hat{t}}_{1 \rightarrow 2}\cdot \mathbf{t}'_{1 \rightarrow 2}}{\mathbf{\hat{t}}_{1 \rightarrow 2}\cdot \mathbf{\hat{t}}_{1 \rightarrow 2}},\\
s_{1\rightarrow 3} = \frac{\mathbf{\hat{t}}_{1 \rightarrow 3}\cdot \mathbf{t}'_{1 \rightarrow 3}}{\mathbf{\hat{t}}_{1 \rightarrow 3}\cdot \mathbf{\hat{t}}_{1 \rightarrow 3}},
\]
where $\mathbf{\hat{t}}_{i \rightarrow j}$ is the TFT predicted translation (up to a scale) from camera $i$ to camera $j$, $\mathbf{t}'_{i \rightarrow j}$ is the marker-calibrated translation in real metric scale. We set the extrinsics of the first camera to be the identity matrix. We use Intel's dynamic calibration tool to calibrate intrinsics.
In cases of repetitive textures, where SuperPoint+SuperGlue were unreliable, we made correspondences manually. 

\subsection{Annotation of Pose Ground-Truths}

Previous works have used ArUco markers to automate pose estimation through Perspective-n-Points, and we do this also. 
Specifically, 
we used ArUco markers to automate the annotation of \textit{still object} poses (i.e. not moving), following LINEMOD~\cite{brachmann2014learning} and NOCS~\cite{wang2019normalized}. We note that this step has two limitations:

\begin{enumerate}
    \item Markers in the scene detract from realism and compromise dataset integrity: training on marker-augmented imagery may result in overfitting to these artificial patterns.
    \item Marker-based annotations are inapplicable to moving objects (since the markers are typically placed on the background), thus limiting the setup to annotating still objects.
\end{enumerate}

We compensate for both of these issues: we remove marker appearances from the dataset (after using them for annotation), and we introduce additional tools to annotate the moving objects. We describe these compensation steps next. 

\begin{figure}[ht]

    \centering
    \includegraphics[width=\linewidth]{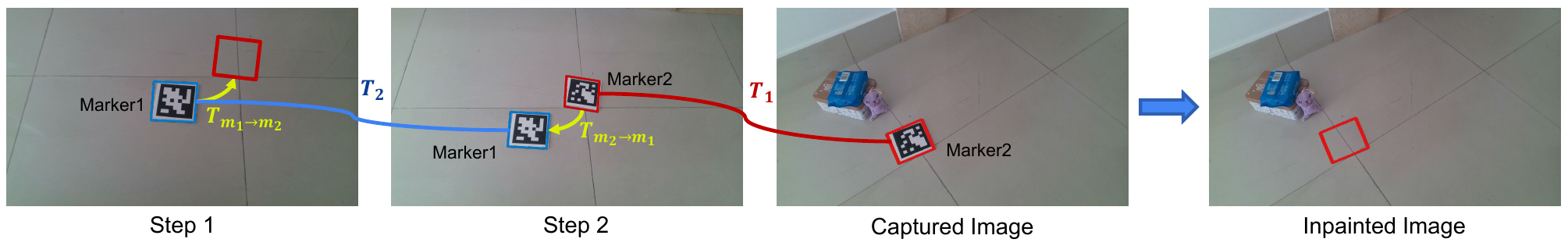}
    \caption{Illustration of the marker inpainting process.}
    \label{fig:inpaint}
    
\end{figure}

\paragraph{Marker removal}
We {removed} the marker appearances from the dataset, 
using a marker inpainting strategy, detailed as follows. In step 1, we place a marker (Marker 1) somewhere within the camera's field of view. We then record a short video with this marker in view. 
In step 2, we place a second marker (Marker 2) at a chosen distance from the first, and record another video. In step 3, we remove Marker 1, and begin the \textit{actual} object capture process (with only Marker 2 present). After this process, we end up with: (1) frames with Marker 1 only, which clearly depict the surface where Marker 2 will later appear; 
(2) frames capturing both markers, providing helpful calibration cues; (3) frames with Marker 2 only, which represent our main capture. 
We leverage Marker 2 for automated pose tracking, and manually correct the tracking every 40 frames in case of drift. 
We use the frames from the first two steps to seamlessly inpaint~\cite{perez2023poisson} Marker 2's area for the final dataset, as depicted in Figure~\ref{fig:inpaint}. 

\paragraph{Annotation of Moving Object Poses}
For moving objects, the traditional marker-based tracking approach is insufficient, as the scene markers do not move in tandem with the objects. In such instances, we harnessed the capabilities of BundleTrack~\cite{wen2021bundletrack}, an advanced RGB image-based tracking algorithm. BundleTrack conducts feature correspondence analysis between successive frames to estimate poses, complemented by a bundle adjustment algorithm to optimize keyframes globally and minimize tracking errors. Despite BundleTrack's proficiency in approximating poses, it is prone to drift, necessitating the manual adjustment of poses every ten frames to ensure precision. We 
adopted a rigorous manual checking process, examining 
every tenth frame for 3D pointcloud alignment and 2D multi-view agreement. 
We detail the annotation tool and its graphical interface in the supplementary. This delicate process represents the most labor-intensive aspect of our annotation pipeline. 

In order to verify that our annotation process gives accurate poses, we conducted an experiment where we used our setup to manually annotate the poses of \textit{Scene 1} from NOCS-REAL275~\cite{wang2019normalized} dataset, and compare the annotations to the provided ground truths. We recorded an average pose annotation error of $0.9^\circ$ for rotation and $2.3$mm for translation.


\paragraph{Generation of Segmentation Masks}
Upon successful pose annotation, we generated occlusion-aware segmentation masks, via depth rendering. 
In cases where hand interactions are involved, we employed the segmentation model SAM~\cite{kirillov2023segment} to estimate hand masks, and subsequently subtracted these from the previously computed object masks to obtain the occlusion-aware segmentation.  


\section{Dataset Statistics}

In pursuit of diversity, we placed the objects in ten disparate environments, each featuring varying levels and configurations of occlusion. We captured 10 video sequences per scene, each sequence including 1 to 7 objects. We captured RGB and depth data using an Intel RealSense D415 camera, with a resolution of $1280\times 720$. The capture process spans distances ranging from 0.5m to 1.5m.

\subsection{Annotation Distribution and Object Diversity}
\begin{figure}[ht]
\centering
    \includegraphics[width=0.9\linewidth]{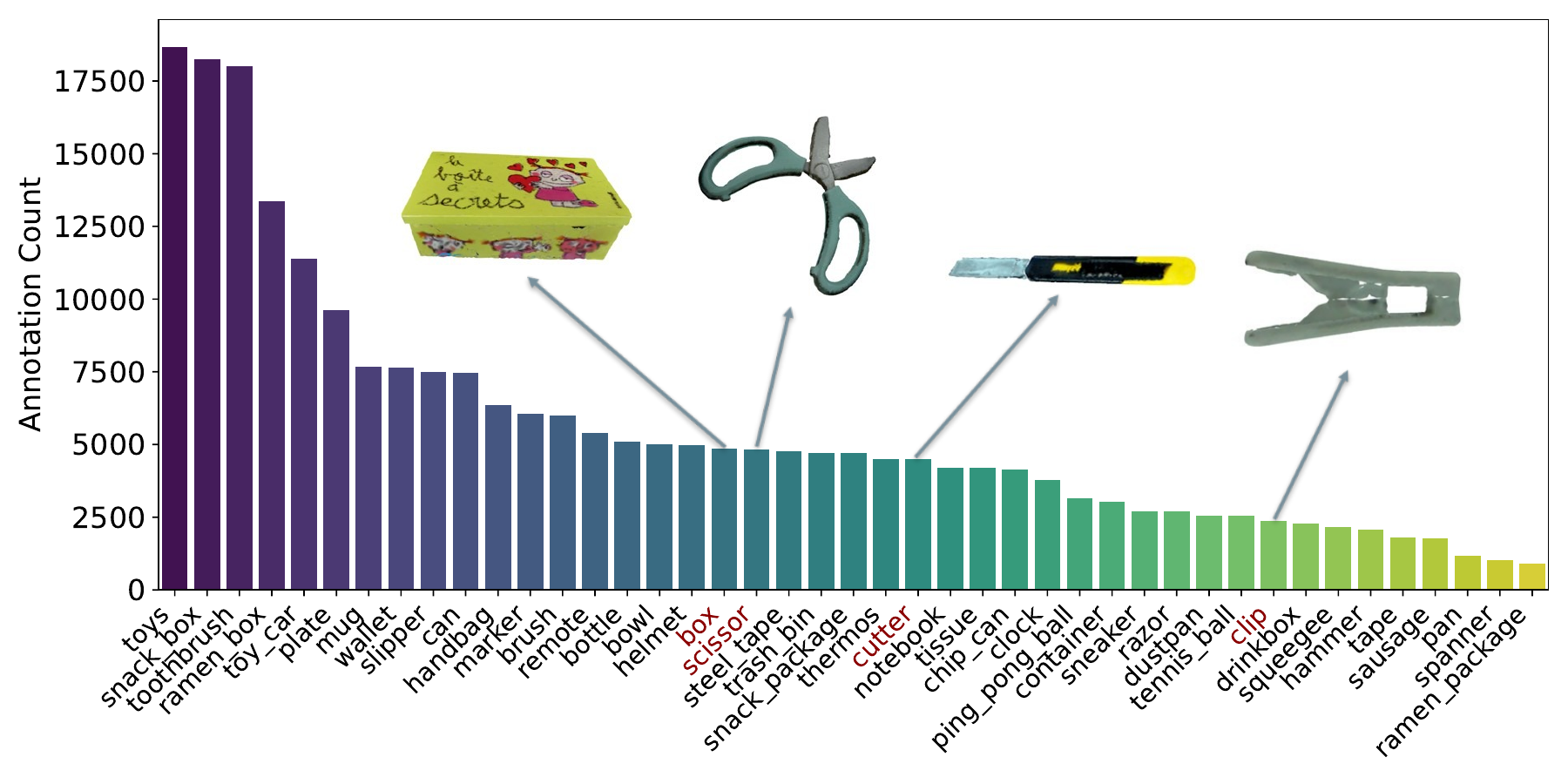}
    
    \caption{Distribution of pose annotation counts across different object categories.
    }
    \label{fig:objdist}
\end{figure}

The comprehensive distribution of pose annotation counts is depicted in Figure~\ref{fig:objdist}, showcasing a diverse array including both rigid and articulated objects. 
As demonstrated in Figure~\ref{fig:sizedist}, our object database encompasses a broad spectrum of object sizes, with the majority measuring near 0.2m along the bounding box diagonal. It also includes larger items such as storage bins, adding to the heterogeneity of the dataset.

%
%


%
\subsection{Variability in Occlusion, Pose, and Environmental Context}
\begin{figure}[ht]
    \centering
    \begin{minipage}{0.58\linewidth}
    \includegraphics[width=\linewidth]{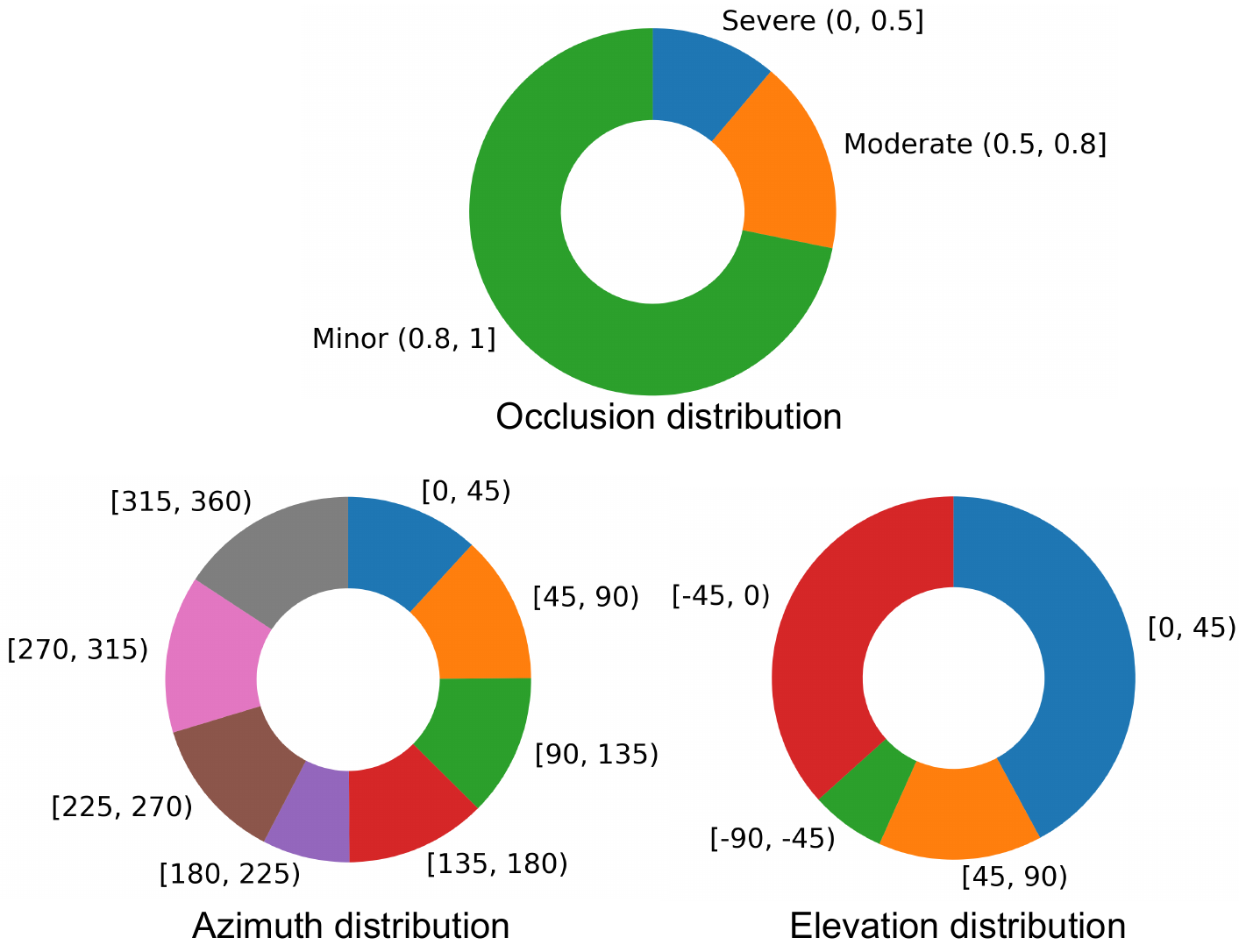}
    \caption{ Distribution of occlusions, azimuth and elevation viewing angles.
    }
    \label{fig:posedist}
\end{minipage}
\begin{minipage}{0.4\linewidth}
	\centering
        \includegraphics[width=\linewidth]{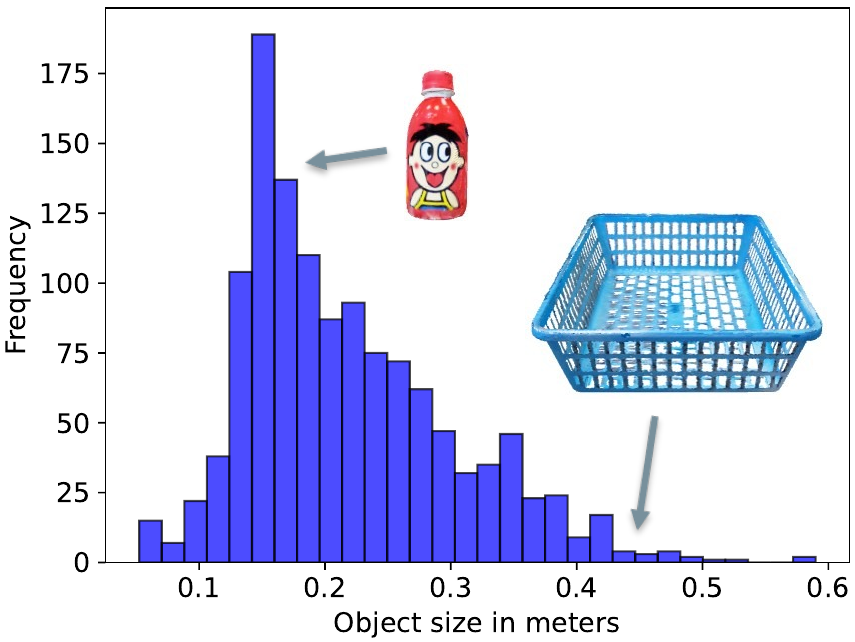}
        
        \caption{Distribution of object sizes. 
        }
        \label{fig:sizedist}
    \end{minipage}

\end{figure}

\begin{figure}[ht]
    \centering
    \subcaptionbox{basket}{\includegraphics[width=0.19\textwidth]{figures/basket.jpg}}
    \hfill
    \subcaptionbox{box top}{\includegraphics[width=0.19\textwidth]{figures/box_top.jpg}}
    \hfill
    \subcaptionbox{cabinet}{\includegraphics[width=0.19\textwidth]{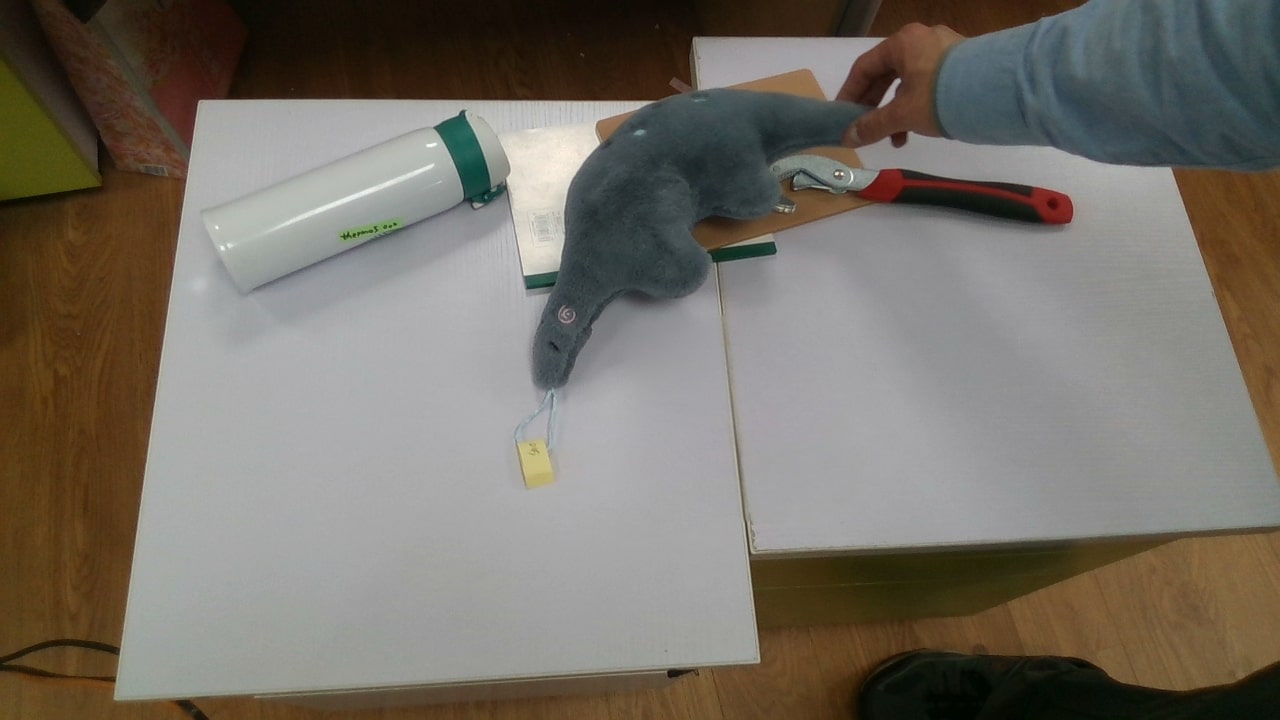}}
    \hfill
    \subcaptionbox{carpet}{\includegraphics[width=0.19\textwidth]{figures/carpet.jpg}}
    \hfill
    \subcaptionbox{chair}{\includegraphics[width=0.19\textwidth]{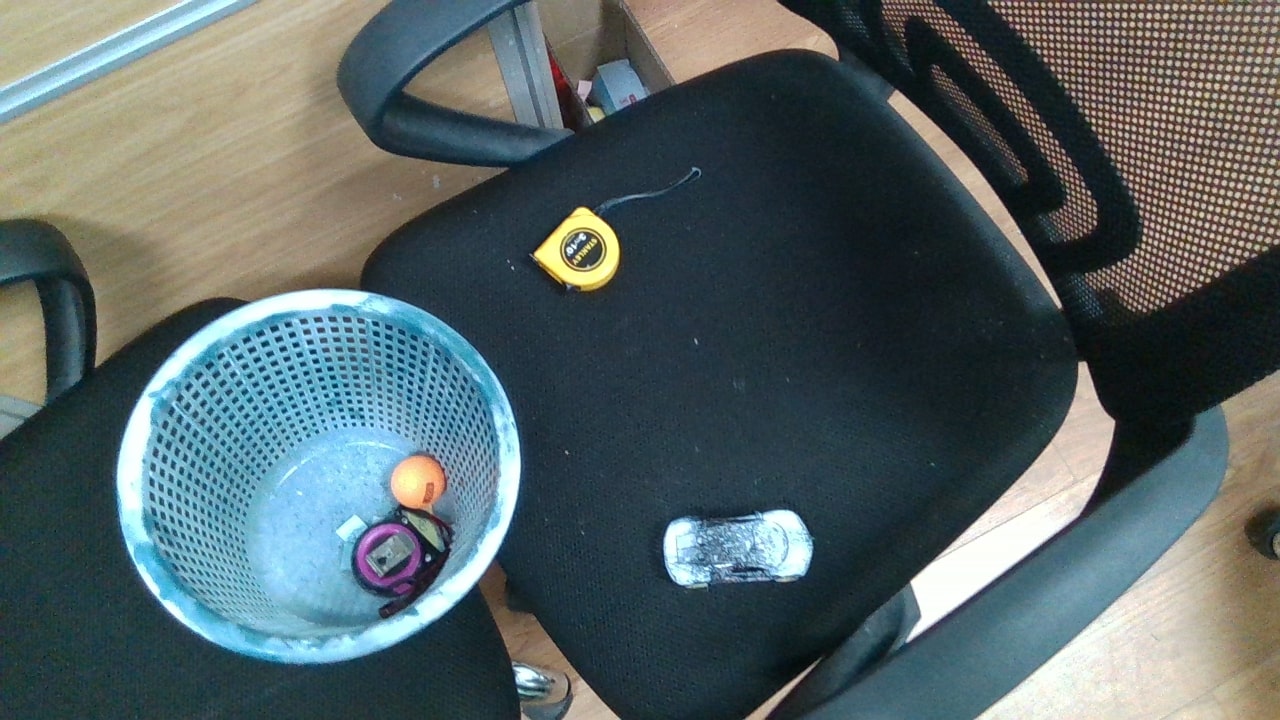}}
    \\
    \subcaptionbox{china floor}{\includegraphics[width=0.19\textwidth]{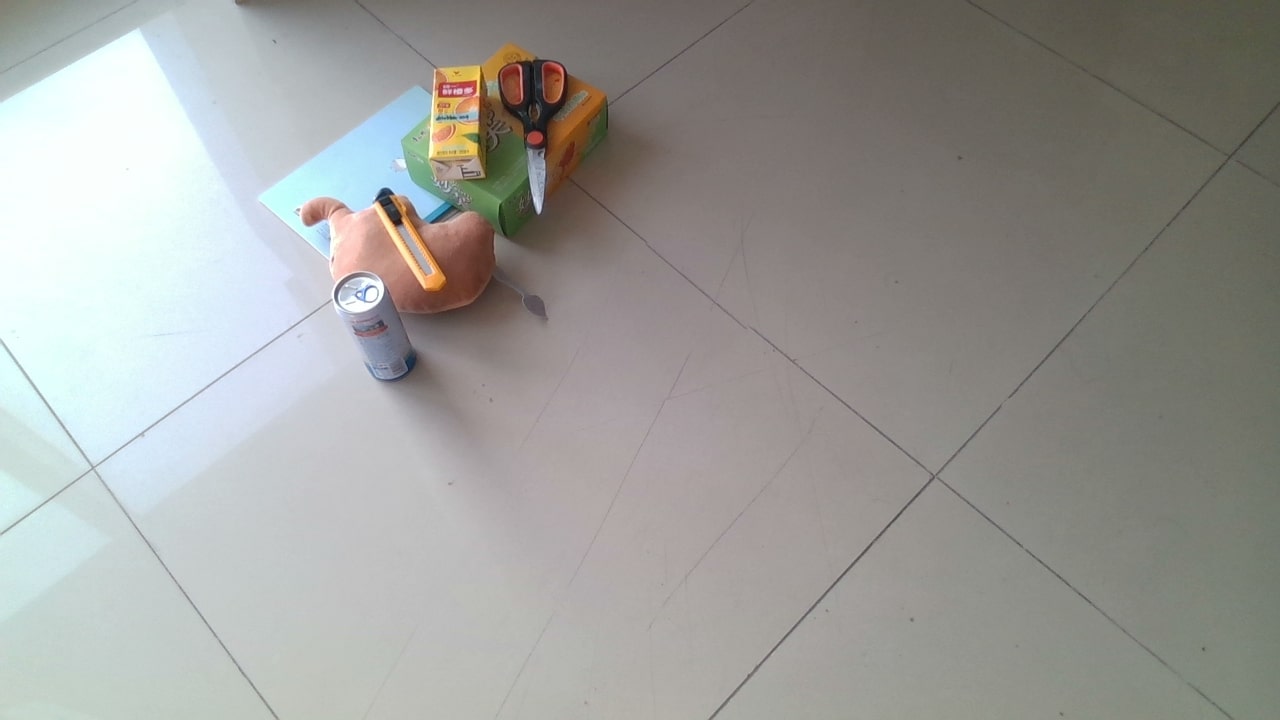}}
    \hfill
    \subcaptionbox{desk1}{\includegraphics[width=0.19\textwidth]{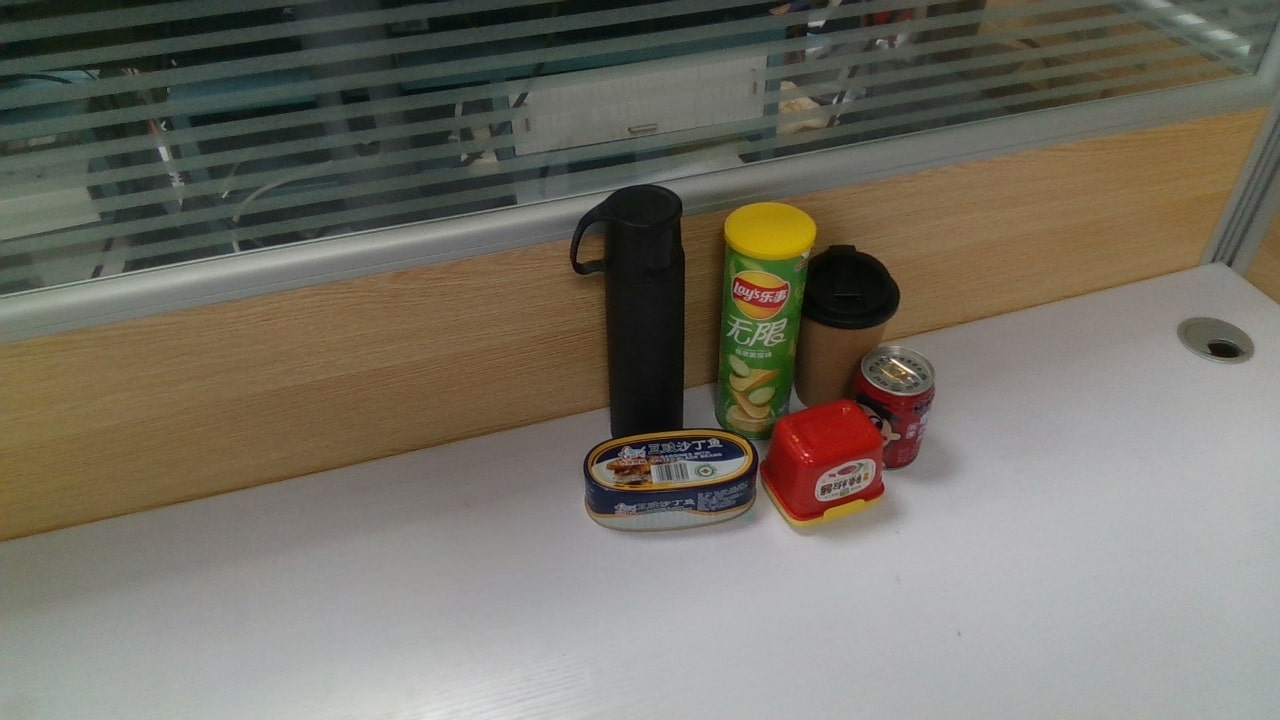}}
    \hfill
    \subcaptionbox{desk2}{\includegraphics[width=0.19\textwidth]{figures/desk2.jpg}}
    \hfill
    \subcaptionbox{stairs}{\includegraphics[width=0.19\textwidth]{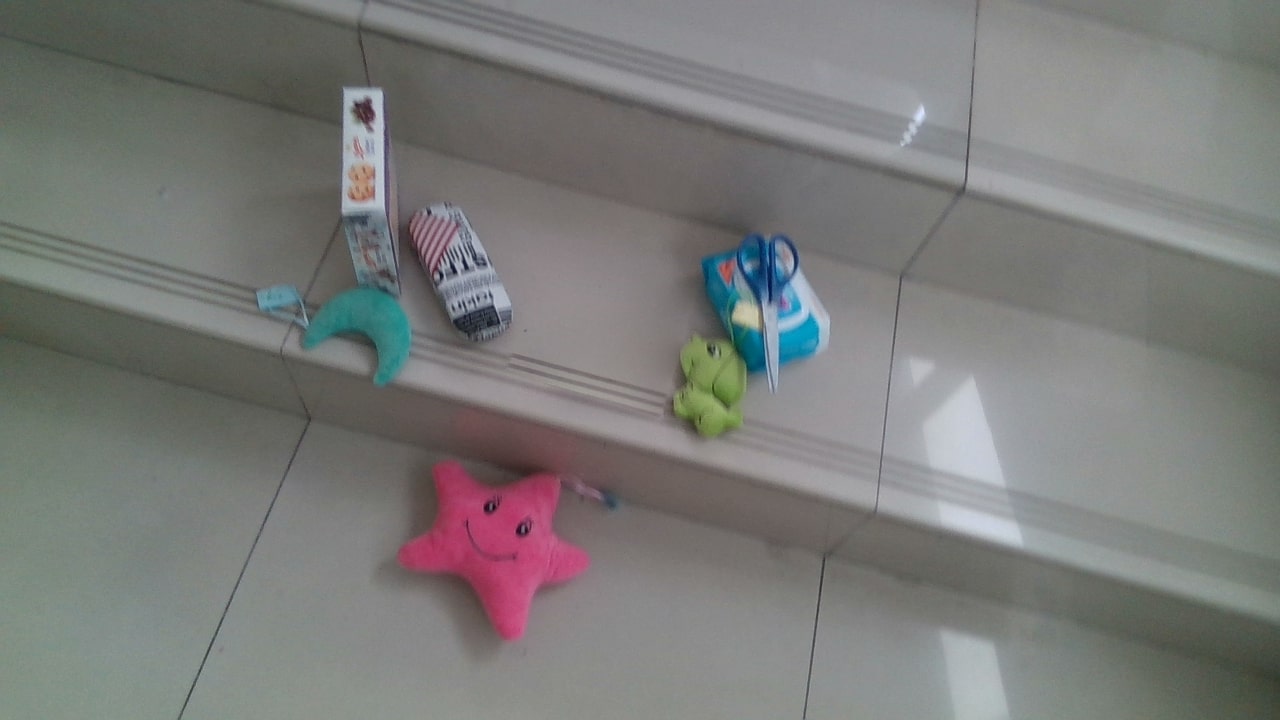}}
    \hfill
    \subcaptionbox{wooden floor}{\includegraphics[width=0.19\textwidth]{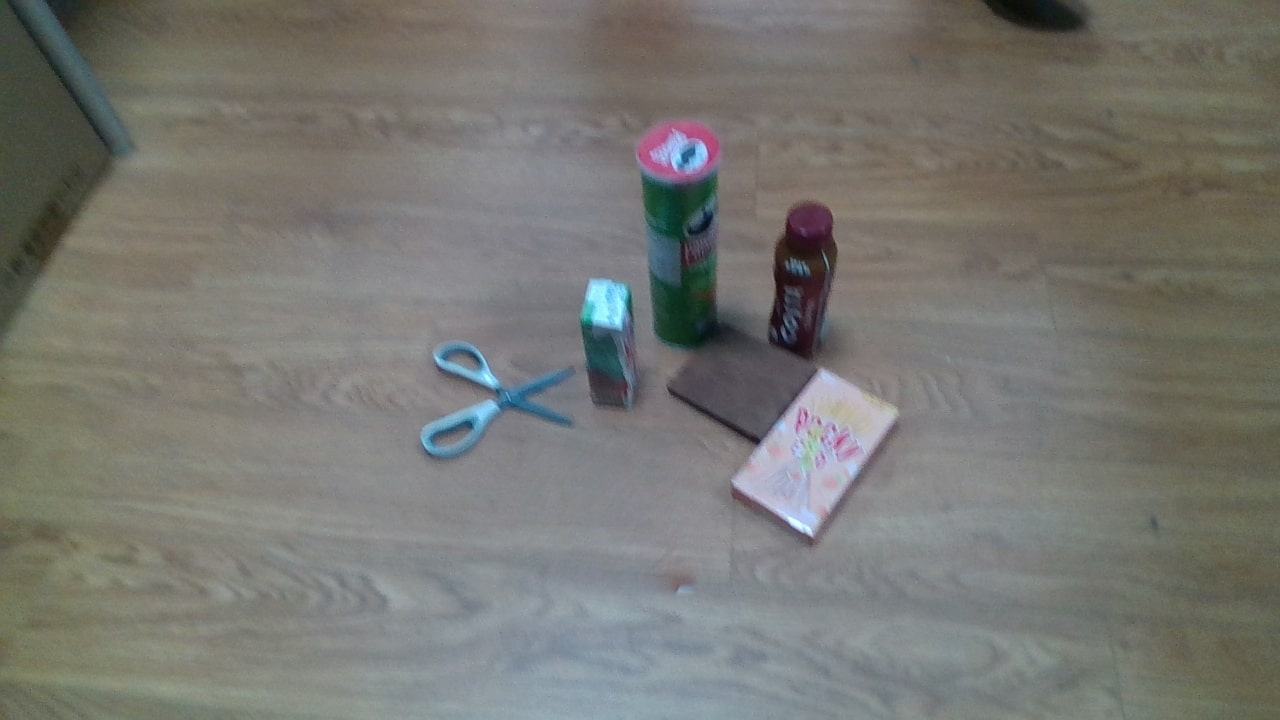}}
    \caption{Sample images from the 10 environments.}
    \label{10-scenes}
\end{figure}

The dataset 
has rich variability in occlusion levels and object poses, as illustrated in Figure~\ref{fig:posedist}, which
outlines the statistical distribution of both visibility ratios and azimuth/elevation angles, indicating a comprehensive coverage reflecting real-world scenarios. These statistics can be helpful for assessing the robustness of pose estimation algorithms. Besides, the ten distinct capturing environments are shown in Figure~\ref{10-scenes}, illustrating diverse backgrounds and lighting conditions.



%
    

%

\subsection{Simulated Training Data and Dataset Split}
\label{sec:datasplit}

To facilitate a comprehensive and equitable evaluation of pose estimation methodologies, we partitioned our entire real-world dataset into validation and test subsets following a 20/80 ratio. The test subset is used for both instance-level and category-level pose estimation evaluation tasks.
The real-world validation subset is used for instance-level pose estimation only. We render another photo-realistic validation subset for category-level pose estimation, in order to cover enough category-level objects that are different from the test set.

Additionally, to support research necessitating extensive training datasets, we generated a large synthetic dataset called \textbf{PACE-Sim} that contains two different training subsets, for instance-level and category-level pose estimation, respectively. We use a physically based renderer~\cite{Denninger2023} to ensure the images are highly photo-realistic. The synthetic training sets consist of 100K images, each featuring dozens of objects, yielding 2.4M annotations.
The distribution of object instances in both PACE and PACE-Sim can be found in supplementary.

    



%

\section{Evaluation Benchmarks}
\label{sec:exp}



We assess both pose estimation and tracking. State-of-the-art (SOTA) pose estimation techniques generally involve two phases: first detecting or segmenting the object, and then estimating its pose within the identified mask. For a fair comparison, we evaluate pose estimation assuming perfect object detection. Detailed SOTA detection performance is available in the supplementary materials but is beyond this dataset's primary scope.


For pose tracking, we provide ground-truth data in the initial frame and require methods to accurately track the object's rotations and translations through the video sequence. Both model-free and model-based trackers are assessed.


\subsection{Pose Estimation Benchmark}
This benchmark is divided into instance-level and category-level pose estimation. The former concerns object instances seen during training (in different scenes), while the latter concerns instances which were not seen but come from the same category. We will also give an in-depth analysis of SOTA's failure modes.

\subsubsection{Instance-Level Pose Estimation}
This task demands the prediction of rotation and translation for known instances from the training set.

\paragraph{Metrics:}
Adhering to the BOP challenge protocol~\cite{hodavn2020bop}, we use Average Recall (AR) of Visible Surface Discrepancy (VSD), Maximum Symmetry-Aware Surface Distance (MSSD), and Maximum Symmetry-Aware Projection Distance (MSPD) as our metrics. Detailed computation of these metrics is described in~\cite{hodavn2020bop}. 
Results are averaged across all instances.

\paragraph{Baselines:}
We assessed four baselines: PPF~\cite{drost2010model}, CosyPose~\cite{labbe2020cosypose}, SurfEmb~\cite{haugaard2022surfemb}, and GDRNPP~\cite{Wang_2021_GDRN}. While PPF does not require training data, the others are SOTA methods dependent on additional training on the PBR dataset. To focus on pose estimation performance, we assume that ground-truth detections are available.

\paragraph{Result Analysis:}
The quantitative analysis is presented in Table~\ref{tab:poseinst}. Notably, while SOTA methods excel on the BOP benchmarks, they do not perform as well as PPF, which relies on local geometric feature matching in point clouds. The relative success of PPF highlights the potential for improvement in SOTA methods, particularly regarding sim-to-real ability and scalability to handle a large set of instances. Qualitative results can be found in the supplementary.

\begin{table*}[ht]

\begin{center}
\resizebox{0.7\linewidth}{!}{
\begin{tabular}{lccccccc}
\toprule
& Modality & Detection & AR$_{VSD}$$\uparrow$ & AR$_{MSSD}$$\uparrow$ & AR$_{MSPD}$$\uparrow$ & AR$\uparrow$ \\
\midrule
PPF~\cite{drost2010model} & D & G.T. & \textbf{53.4} & \textbf{48.1} & \textbf{55.2} & \textbf{52.2} \\
CosyPose~\cite{labbe2020cosypose} & RGB & G.T. & 1.4 & 0.3 & 11.5 & 4.4 \\
SurfEmb~\cite{haugaard2022surfemb} & RGB & G.T. & 6.2 & 3.0 & 17.8 & 9.0  \\
GDRNPP~\cite{Wang_2021_GDRN} & RGB-D & G.T. & 3.6 & 2.1 & 15.4 & 7.0 \\
\bottomrule
\end{tabular}
}
\end{center}

\caption{\textbf{Instance-level pose estimation results} showcasing the robustness of PPF and the potential for improvement in SOTA deep learning-based methods.}
\label{tab:poseinst}

\end{table*}

\begin{table*}[ht]

\begin{center}
\resizebox{\linewidth}{!}{
\begin{tabular}{lccccccccc}
\toprule
\multirow{3}*{} & \multirow{3}*{Detection} & 
\multirowcell{3}{IoU$_{25}$$\uparrow$} & \multirowcell{3}{IoU$_{50}$$\uparrow$} & \multicolumn{6}{c}{AP}\\
\cmidrule(lr){5-10}
~ & ~ &  & & \multirowcell{2}{0:20$^\circ$$\uparrow$} & \multirowcell{2}{0:60$^\circ$$\uparrow$} & \multirowcell{2}{0:5cm$\uparrow$} & \multirowcell{2}{0:15cm$\uparrow$} & \multirowcell{2}{0:20$^\circ$\\ 0:5cm}\multirowcell{2}{$\Big\uparrow$} & \multirowcell{2}{0:60$^\circ$\\ 0:15cm}\multirowcell{2}{$\Big\uparrow$} \\
& & & & & & \\
\midrule
NOCS~\cite{wang2019normalized} &  Mask-RCNN & 0.2/0.0 & 0.0/0.0 & 1.2/0.0 & 4.3/0.0 & 17.0/0.0 & 33.5/0.0 & 1.1/0.0 & 4.2/0.0 \\
HS-Pose~\cite{zheng2023hs} & G.T.  & 36.6/0.0 & 2.7/0.0 & 5.3/0.0 & 8.6/0.3 & \textbf{48.7}/34.1 & \textbf{81.0}/\textbf{75.6} & 3.9/0.0 & 8.1/0.3 \\
SGPA~\cite{chen2021sgpa} &  G.T.  & 2.6/0.0 & 1.2/0.0 & 5.6/1.1 & 11.4/\textbf{7.3} & 16.1/6.8 & 50.7/17.5 & 3.3/0.9 & 10.1/\textbf{7.2} \\
DualPoseNet~\cite{lin2021dualposenet} &  G.T. & 24.2/0.3 & 0.1/0.0 & 5.5/0.0 & 8.1/0.0 & 18.6/27.3 & 63.4/65.2 & 1.8/0.0 & 6.5/0.0 \\
SAR-Net~\cite{lin2022sar} &  G.T.  & 22.8/0.1 & 0.2/0.0 & 5.3/0.0 & 8.8/0.8 & 48.6/\textbf{38.5} & 79.8/69.7 & 3.7/0.0 & 8.2/0.8 \\
CPPF++~\cite{you2022cppf++} &  G.T.  & \textbf{44.5}/\textbf{3.6} & \textbf{4.4}/0.0 & \textbf{15.2}/\textbf{1.7} & \textbf{27.3}/6.2 & 35.3/31.1 & 74.0/66.7 & \textbf{9.9}/\textbf{1.1} & \textbf{24.9}/5.9 \\
ANCSH~\cite{li2019category} &  G.T.  & -/0.0 & -/0.0 & -/0.0 & -/0.2 & -/18.6 & -/50.4 & -/0.0 & -/0.2 \\
\bottomrule
\end{tabular}
}
\end{center}

\caption{\textbf{Category-level pose estimation benchmark} combining the performance metrics for both rigid and articulated object pose estimation, separated by slash.}
\label{tab:merged_nocs}

\end{table*}

\subsubsection{Category-Level Pose Estimation}
Category-level pose estimation tasks involve predicting the 3D bounding box dimensions, rotation, and translation of target instances, given only the category information a priori.

\paragraph{Metrics:}
We adopted from NOCS~\cite{wang2019normalized}, calculating mean Average Precision (AP) across predefined angle and translation thresholds. Specifically, AP@0:20$^\circ$ represents AP averaged from 0$^\circ$ to 20$^\circ$ at 1$^\circ$ intervals; AP@0:5cm is averaged from 0cm to 5cm at 0.25cm intervals; AP@0:20$^\circ$,0:5cm combines these angle and translation thresholds. IoU$_{25}$ and IoU$_{50}$ denote AP for 3D bounding box matches at IoU thresholds of 25 and 50, respectively. 
Results are separately reported for both rigid and articulated objects.

\paragraph{Baselines:}
We evaluated seven recent category-level pose estimation methods: NOCS~\cite{wang2019normalized}, HS-Pose~\cite{zheng2023hs}, SGPA~\cite{chen2021sgpa}, DualPoseNet~\cite{lin2021dualposenet}, SAR-Net~\cite{lin2022sar}, CPPF++~\cite{you2022cppf++} and ANCSH~\cite{li2019category}, the latter specifically designed for articulated objects. For adapting rigid-object-focused methods to articulated objects, each movable part is considered a distinct category. 
Ground-truth detections are presumed for each method, except for NOCS, which outputs both instance masks and pose estimations using a unified network. 

\paragraph{Result Analysis:}
Quantitative comparisons are listed in 
Table~\ref{tab:merged_nocs}. On rigid objects, HS-Pose and SAR-Net show fair performance on AP metrics for translation, but falls short on rotation AP metrics compared to CPPF++, suggesting a proficiency in size and translation prediction but not rotation. CPPF++ shows the best performance but is weaker on translation estimation. 
For articulated objects, all methods face challenges due to the movable parts, with ANCSH underperforming on the large-scale real-world dataset, indicative of a significant sim-to-real gap. NOCS shows instability, likely due to its inaccurate mask prediction. Qualitative results are available in supplementary.

\subsubsection{Failure Mode Analysis}
This section delves into the analysis of two critical questions that underline the failure of SOTA deep learning-based methods in pose estimation: 1) Is the failure attributed to the sim-to-real gap, considering the exclusive use of synthetic RGB-D images for training? 2) Does the failure stem from the inherent large-scale complexity of the PACE dataset, which poses a significant challenge for baseline models?

\paragraph{Ablation Setup:} To explore these questions, we introduced a new 80/10/10 train/validation/test split for the real-world data, diverging from the 0/20/80 split described in Section~\ref{sec:datasplit}. We then examined the following settings:

\begin{itemize}
\item \textit{Real2Real Single Instance/Category Fitting:} This basic setup involves training on a single instance or category selected (we randomly select the category \textit{can} and sample a can with object ID 57) from the 80/10/10 split of real data.
\item \textit{Sim2Real Single Instance/Category Fitting:} This scenario differs from the first by utilizing all simulated data for training and validation, with evaluation on the same real-world test split.
\item \textit{Real2Real All Instance/Category Fitting:} Unlike the first setting, this involves training, validation, and testing across all instances or categories.
\end{itemize}

\begin{table}[!ht]
    \centering
    \resizebox{\linewidth}{!}{
    \begin{tabular}{lccccccccc}
        \toprule
        \multirow{2}*{Training Setting} & \multicolumn{3}{c}{Instance-Level (AR\%)$\uparrow$} & \multicolumn{6}{c}{Category-Level (mean AP\%@0:60$^\circ$, 0:15cm)$\uparrow$}\\
        \cmidrule(lr){2-4}\cmidrule(lr){5-10}
         & CosyPose & SurfEmb & GDRNPP & NOCS & HS-Pose & SGPA & DualPoseNet & SAR-Net & CPPF++\\
        \midrule
        Real2Real Single Inst./Cat. & \textbf{76.4} & \textbf{86.7} & \textbf{88.2} & \textbf{76.9} & \textbf{81.2} & \textbf{86.7} & \textbf{58.6} & \textbf{59.3} & \textbf{83.4} \\ 
        \midrule
        
        Sim2Real Single Inst./Cat. & 67.1 (\textcolor{BrickRed}{-9.3$\downarrow$}) & 73.7 (\textcolor{BrickRed}{-13.0$\downarrow$}) & 78.1 (\textcolor{BrickRed}{-10.1$\downarrow$}) & 55.8 (\textcolor{BrickRed}{-21.1$\downarrow$}) & 1.1 (\textcolor{BrickRed}{-80.1$\downarrow$}) & 0.3 (\textcolor{BrickRed}{-86.4$\downarrow$}) & 1.4 (\textcolor{BrickRed}{-57.2$\downarrow$}) & 33.2 (\textcolor{BrickRed}{-26.1$\downarrow$}) & 55.8 (\textcolor{BrickRed}{-27.6$\downarrow$})\\
        
        Real2Real All Inst./Cat. & 8.2 (\textcolor{BrickRed}{-68.2$\downarrow$})& 11.2 (\textcolor{BrickRed}{-75.5$\downarrow$})& 9.7 (\textcolor{BrickRed}{-78.5$\downarrow$})& 0.3 (\textcolor{BrickRed}{-76.6$\downarrow$})& 5.6 (\textcolor{BrickRed}{-75.6$\downarrow$})& 9.8 (\textcolor{BrickRed}{-76.9$\downarrow$})& 1.2 (\textcolor{BrickRed}{-57.4$\downarrow$})& 2.1 (\textcolor{BrickRed}{-57.2$\downarrow$}) & 13.7 (\textcolor{BrickRed}{-69.7$\downarrow$}) \\
        
        \bottomrule
    \end{tabular}
    }
    
    \caption{\textbf{Ablation Study:} This study illustrates the challenges models face when transitioning from real-to-real settings. It showcases performance disparities when scaling from single instances/categories to the entire dataset, emphasizing the scalability challenge from PACE and the profound impact of sim-to-real discrepancies on depth-dependent methodologies.}
    \label{tab:falure_pose}
    
\end{table}

\paragraph{Result Analysis and Discussions:}
Our analysis in Table~\ref{tab:falure_pose} reveals that baseline models effectively handle single instance or category fitting under the \textit{Real2Real Single Instance/Category Fitting} setting, provided the training and test distributions align. However, the sim-to-real gap, especially noticeable in depth-input methods like DualPoseNet, HS-Pose, and SGPA, highlights significant challenges, as these methods struggle with the depth noise in real world.

Additionally, learning across all instances/categories leads to a notable performance drop, as shown in Table~\ref{tab:falure_pose}. This drop, together with suboptimal training data performance and loss plateauing, suggests underfitting and highlights the complexity posed by the PACE dataset's diversity. Innovative approaches or larger foundational models might be necessary to address these challenges and effectively bridge the sim-to-real gap.


\subsection{Pose Tracking Benchmark}
We distinguish SOTA pose tracking techniques into model-based, utilizing a 3D CAD model, and model-free, which only require an initial pose from first frame.


\paragraph{Metrics:} 


For model-based tracking, we measure the area under curve (AUC) for ADD, ADD-S \cite{xiang2017posecnn}, and ADD(-S) metrics, with a maximum threshold of 0.1m \cite{wang2019densefusion}. The ADD metric, introduced in \cite{hinterstoisser2013model}, computes the average per-point distance between point clouds transformed by the predicted and ground-truth poses. ADD-S accounts for point correspondence ambiguities in symmetric objects. 

For model-free tracking, we adopt four metrics from prior studies \cite{wang20206}: 1) $\mathbf{5^\circ5cm}$, indicating the percentage of predictions within 5° rotation and 5cm translation errors. 2) \textbf{IoU25}, the percentage of instances where the intersection over union of the 3D bounding boxes exceeds $25\%$. 3) $\mathbf{R_{err}}$ and 4) $\mathbf{T_{err}}$, the mean rotational and translational errors in degrees and centimeters, respectively, only considering instances with IoU$>$25\%. We exclude objects with visibility under 10\% as per the BOP convention \cite{hodavn2020bop}. Results are averaged separately across rigid and articulated object categories.

\begin{table}[ht]

\begin{center}
\resizebox{0.57\linewidth}{!}{
\begin{tabular}{lcccc}
\toprule
& Modality & ADD$\uparrow$ & ADD-S$\uparrow$ & ADD(-S)$\uparrow$ \\
\midrule
RBOT~\cite{tjaden2018region}  & RGB    & 7.1/0.5 & 10.3/0.8 & 7.4/0.5  \\
ICG~\cite{stoiber2022iterative}   & RGB-D  & \textbf{35.6}/\textbf{10.1} & \textbf{48.1}/\textbf{15.2} & \textbf{38.1}/\textbf{10.1}  \\
\bottomrule
\end{tabular}
}
\end{center}

\caption{\textbf{Model-Based Pose Tracking.} We report the area under curve (AUC) with respect to ADD, ADD-S and ADD(-S). The higher value, the better performance.
}

\label{tab:track1}
\end{table}

\begin{table*}[h]
\begin{center}
\resizebox{0.8\linewidth}{!}{
\begin{tabular}{lcccccc}
\toprule
& Training-Free & Modality &  5$^\circ$5cm$\uparrow$ & IoU25$\uparrow$ & R$_{err}$$\downarrow$ & T$_{err}$$\downarrow$  \\
\midrule
BundleTrack~\cite{wen2021bundletrack}  & \cmark & RGB      & 6.4/\textbf{11.2}  & 9.1/14.1 & \textbf{3.2}/\textbf{5.5}  & 2.6/\textbf{0.8}   \\
CAPTRA~\cite{weng2021captra}    &  \xmark & D       & \textbf{12.9}/4.4  & \textbf{45.8}/\textbf{20.6}  & 19.2/40.9  & 2.2/1.5 \\
6-PACK~\cite{wang20206}    & \xmark  & RGB-D    & 9.2/3.9  & 23.1/16.7  & 17.7/33.6  & \textbf{2.1}/1.2 \\
\bottomrule
\end{tabular}
}
\end{center}

\caption{\textbf{Model-Free Pose Tracking}. Both rigid and articulated results are reported. For 5$^\circ$5cm and IoU25, the higher value means better performance while for R$_{err}$ and T$_{err}$, the situation is reversed. Note that R$_{err}$ and T$_{err}$ are respect to IoU25 since objects with IoU$\leq 25\%$ are not counted. 
}

\label{tab:track2}
\end{table*}

\paragraph{Baselines:} For tracking methods using object masks, we applied the ground-truth masks directly. Except for CAPTRA, which also predicts the 3D bounding box size, all methods estimate only rotation and translation; however, we used ground-truth bounding box sizes for consistency in evaluations. Unlike other methods that treat each part of an articulated object independently, CAPTRA considers the entire object as one entity, trained on our synthetic PBR data.

\paragraph{Result and Failure Mode Analysis:}
Tables \ref{tab:track1} and \ref{tab:track2} display the performance of state-of-the-art (SOTA) tracking methods on the PACE dataset, highlighting both rigid and articulated object results. The highest achieved ADD(-S) for model-based methods is only 38.1\% for rigid and 10.1\% for articulated objects by ICG. Furthermore, Table \ref{tab:track2} reveals severe limitations of model-free methods, with the best 5$^\circ$5cm accuracy under 13\%, posing major tracking challenges.

The convention followed by 6-PACK~\cite{wang20206} and BundleTrack~\cite{wen2021bundletrack} disregards objects with IoU$\leq 25\%$ when computing rotational and translational errors. However, this approach could overestimate the performance on R$_{err}$ and T$_{err}$. BundleTrack~\cite{wen2021bundletrack} reports lower performance in 5$^\circ$5cm and IoU25 metrics yet shows seemingly better results for R$_{err}$ and T$_{err}$, suggesting that it can only track well within a threshold but cannot recover once the object is lost. 
Qualitative results can be found in the supplementary.





\section{Conclusions and Limitations}

We present PACE, a new large-scale benchmark for 3D object pose estimation and tracking, designed with various occlusion levels across ten diverse environments. This benchmark both reflects prior progress and urges further research to tackle real-world complexities. Our analysis indicates significant advances in pose estimation, yet underscores a notable real-world performance gap, particularly with articulated objects and in terms of robustness and scalability. Future efforts could focus on advanced sim-to-real techniques and developing larger models that capture a broader spectrum of object features and environments.

\paragraph{Limitations}
Due to resource limitations, PACE lacks large and valuable objects like tables, cameras, and laptops, which may affect its breadth in representing real-world object scenarios.
%
%

\section{Acknowledgements}
This work is supported by the National Key Research and Development Project of China (No. 2022ZD0160102),
National Key Research and Development Project of China (No. 2021ZD0110704), Shanghai Artificial Intelligence Laboratory, XPLORER PRIZE grants. Yang You is also supported in part by the Outstanding Doctoral Graduates Development Scholarship of Shanghai Jiao Tong University.

%
%
\bibliographystyle{splncs04}
\bibliography{main}

\begin{thebibliography}{10}
\providecommand{\url}[1]{\texttt{#1}}
\providecommand{\urlprefix}{URL }
\providecommand{\doi}[1]{https://doi.org/#1}

\bibitem{ahmadyan2021objectron}
Ahmadyan, A., Zhang, L., Ablavatski, A., Wei, J., Grundmann, M.: Objectron: A large scale dataset of object-centric videos in the wild with pose annotations. In: Proceedings of the IEEE/CVF conference on computer vision and pattern recognition. pp. 7822--7831 (2021)

\bibitem{avetisyan2019scan2cad}
Avetisyan, A., Dahnert, M., Dai, A., Savva, M., Chang, A.X., Nie{\ss}ner, M.: Scan2cad: Learning cad model alignment in rgb-d scans. In: Proceedings of the IEEE/CVF Conference on computer vision and pattern recognition. pp. 2614--2623 (2019)

\bibitem{brachmann2014learning}
Brachmann, E., Krull, A., Michel, F., Gumhold, S., Shotton, J., Rother, C.: Learning 6d object pose estimation using 3d object coordinates. In: Computer Vision--ECCV 2014: 13th European Conference, Zurich, Switzerland, September 6-12, 2014, Proceedings, Part II 13. pp. 536--551. Springer (2014)

\bibitem{chen2021sgpa}
Chen, K., Dou, Q.: Sgpa: Structure-guided prior adaptation for category-level 6d object pose estimation. In: Proceedings of the IEEE/CVF International Conference on Computer Vision. pp. 2773--2782 (2021)

\bibitem{deng2021poserbpf}
Deng, X., Mousavian, A., Xiang, Y., Xia, F., Bretl, T., Fox, D.: Poserbpf: A rao--blackwellized particle filter for 6-d object pose tracking. IEEE Transactions on Robotics  \textbf{37}(5),  1328--1342 (2021)

\bibitem{Denninger2023}
Denninger, M., Winkelbauer, D., Sundermeyer, M., Boerdijk, W., Knauer, M., Strobl, K.H., Humt, M., Triebel, R.: Blenderproc2: A procedural pipeline for photorealistic rendering. Journal of Open Source Software  \textbf{8}(82), ~4901 (2023). \doi{10.21105/joss.04901}, \url{https://doi.org/10.21105/joss.04901}

\bibitem{detone2018superpoint}
DeTone, D., Malisiewicz, T., Rabinovich, A.: Superpoint: Self-supervised interest point detection and description. In: Proceedings of the IEEE conference on computer vision and pattern recognition workshops. pp. 224--236 (2018)

\bibitem{drost2010model}
Drost, B., Ulrich, M., Navab, N., Ilic, S.: Model globally, match locally: Efficient and robust 3d object recognition. In: 2010 IEEE computer society conference on computer vision and pattern recognition. pp. 998--1005. Ieee (2010)

\bibitem{garon2017deep}
Garon, M., Lalonde, J.F.: Deep 6-dof tracking. IEEE transactions on visualization and computer graphics  \textbf{23}(11),  2410--2418 (2017)

\bibitem{garrido2014automatic}
Garrido-Jurado, S., Mu{\~n}oz-Salinas, R., Madrid-Cuevas, F.J., Mar{\'\i}n-Jim{\'e}nez, M.J.: Automatic generation and detection of highly reliable fiducial markers under occlusion. Pattern Recognition  \textbf{47}(6),  2280--2292 (2014)

\bibitem{guo2023handal}
Guo, A., Wen, B., Yuan, J., OTHERS: Handal: A dataset of real-world manipulable object categories with pose annotations, affordances, and reconstructions  (2023), \url{https://arxiv.org/pdf/2308.01477.pdf}

\bibitem{haugaard2022surfemb}
Haugaard, R.L., Buch, A.G.: Surfemb: Dense and continuous correspondence distributions for object pose estimation with learnt surface embeddings. In: Proceedings of the IEEE/CVF Conference on Computer Vision and Pattern Recognition. pp. 6749--6758 (2022)

\bibitem{hinterstoisser2013model}
Hinterstoisser, S., Lepetit, V., Ilic, S., Holzer, S., Bradski, G., Konolige, K., Navab, N.: Model based training, detection and pose estimation of texture-less 3d objects in heavily cluttered scenes. In: Computer Vision--ACCV 2012: 11th Asian Conference on Computer Vision, Daejeon, Korea, November 5-9, 2012, Revised Selected Papers, Part I 11. pp. 548--562. Springer (2013)

\bibitem{hodavn2020bop}
Hoda{\v{n}}, T., Sundermeyer, M., Drost, B., Labb{\'e}, Y., Brachmann, E., Michel, F., Rother, C., Matas, J.: Bop challenge 2020 on 6d object localization. In: Computer Vision--ECCV 2020 Workshops: Glasgow, UK, August 23--28, 2020, Proceedings, Part II 16. pp. 577--594. Springer (2020)

\bibitem{jampani2023navi}
Jampani, V., Maninis, K.K., Engelhardt, A., Karpur, A., Truong, K., Sargent, K., Popov, S., Araujo, A., Martin-Brualla, R., Patel, K., et~al.: Navi: Category-agnostic image collections with high-quality 3d shape and pose annotations. arXiv preprint arXiv:2306.09109  (2023)

\bibitem{julia2018critical}
Juli{\`a}, L.F., Monasse, P.: A critical review of the trifocal tensor estimation. In: Image and Video Technology: 8th Pacific-Rim Symposium, PSIVT 2017, Wuhan, China, November 20-24, 2017, Revised Selected Papers 8. pp. 337--349. Springer (2018)

\bibitem{jung2024housecat6d}
Jung, H., Wu, S.C., Ruhkamp, P., Zhai, G., Schieber, H., Rizzoli, G., Wang, P., Zhao, H., Garattoni, L., Meier, S., et~al.: Housecat6d-a large-scale multi-modal category level 6d object perception dataset with household objects in realistic scenarios. In: Proceedings of the IEEE/CVF Conference on Computer Vision and Pattern Recognition. pp. 22498--22508 (2024)

\bibitem{kirillov2023segment}
Kirillov, A., Mintun, E., Ravi, N., Mao, H., Rolland, C., Gustafson, L., Xiao, T., Whitehead, S., Berg, A.C., Lo, W.Y., et~al.: Segment anything. arXiv preprint arXiv:2304.02643  (2023)

\bibitem{labbe2020cosypose}
Labb{\'e}, Y., Carpentier, J., Aubry, M., Sivic, J.: Cosypose: Consistent multi-view multi-object 6d pose estimation. In: Computer Vision--ECCV 2020: 16th European Conference, Glasgow, UK, August 23--28, 2020, Proceedings, Part XVII 16. pp. 574--591. Springer (2020)

\bibitem{li2019category}
Li, X., Wang, H., Yi, L., Guibas, L., Abbott, A.L., Song, S.: Category-level articulated object pose estimation. Proceedings of the IEEE Conference on Computer Vision and Pattern Recognition  (2020)

\bibitem{lin2022sar}
Lin, H., Liu, Z., Cheang, C., Fu, Y., Guo, G., Xue, X.: Sar-net: Shape alignment and recovery network for category-level 6d object pose and size estimation. In: Proceedings of the IEEE/CVF Conference on Computer Vision and Pattern Recognition. pp. 6707--6717 (2022)

\bibitem{lin2021dualposenet}
Lin, J., Wei, Z., Li, Z., Xu, S., Jia, K., Li, Y.: Dualposenet: Category-level 6d object pose and size estimation using dual pose network with refined learning of pose consistency. In: Proceedings of the IEEE/CVF International Conference on Computer Vision. pp. 3560--3569 (2021)

\bibitem{liu2022akb}
Liu, L., Xu, W., Fu, H., Qian, S., Yu, Q., Han, Y., Lu, C.: Akb-48: A real-world articulated object knowledge base. In: Proceedings of the IEEE/CVF Conference on Computer Vision and Pattern Recognition. pp. 14809--14818 (2022)

\bibitem{liu2021stereobj}
Liu, X., Iwase, S., Kitani, K.M.: Stereobj-1m: Large-scale stereo image dataset for 6d object pose estimation. In: Proceedings of the IEEE/CVF international conference on computer vision. pp. 10870--10879 (2021)

\bibitem{perez2023poisson}
P{\'e}rez, P., Gangnet, M., Blake, A.: Poisson image editing. In: Seminal Graphics Papers: Pushing the Boundaries, Volume 2, pp. 577--582 (2023)

\bibitem{sarlin2020superglue}
Sarlin, P.E., DeTone, D., Malisiewicz, T., Rabinovich, A.: Superglue: Learning feature matching with graph neural networks. In: Proceedings of the IEEE/CVF conference on computer vision and pattern recognition. pp. 4938--4947 (2020)

\bibitem{stoiber2023fusing}
Stoiber, M., Elsayed, M., Reichert, A.E., Steidle, F., Lee, D., Triebel, R.: Fusing visual appearance and geometry for multi-modality 6dof object tracking. arXiv preprint arXiv:2302.11458  (2023)

\bibitem{stoiber2022iterative}
Stoiber, M., Sundermeyer, M., Triebel, R.: Iterative corresponding geometry: Fusing region and depth for highly efficient 3d tracking of textureless objects. In: Proceedings of the IEEE/CVF Conference on Computer Vision and Pattern Recognition. pp. 6855--6865 (2022)

\bibitem{sun2018pix3d}
Sun, X., Wu, J., Zhang, X., Zhang, Z., Zhang, C., Xue, T., Tenenbaum, J.B., Freeman, W.T.: Pix3d: Dataset and methods for single-image 3d shape modeling. In: Proceedings of the IEEE conference on computer vision and pattern recognition. pp. 2974--2983 (2018)

\bibitem{tjaden2018region}
Tjaden, H., Schwanecke, U., Sch{\"o}mer, E., Cremers, D.: A region-based gauss-newton approach to real-time monocular multiple object tracking. IEEE transactions on pattern analysis and machine intelligence  \textbf{41}(8),  1797--1812 (2018)

\bibitem{wang20206}
Wang, C., Mart{\'\i}n-Mart{\'\i}n, R., Xu, D., Lv, J., Lu, C., Fei-Fei, L., Savarese, S., Zhu, Y.: 6-pack: Category-level 6d pose tracker with anchor-based keypoints. In: 2020 IEEE International Conference on Robotics and Automation (ICRA). pp. 10059--10066. IEEE (2020)

\bibitem{wang2019densefusion}
Wang, C., Xu, D., Zhu, Y., Mart{\'\i}n-Mart{\'\i}n, R., Lu, C., Fei-Fei, L., Savarese, S.: Densefusion: 6d object pose estimation by iterative dense fusion. In: Proceedings of the IEEE/CVF conference on computer vision and pattern recognition. pp. 3343--3352 (2019)

\bibitem{Wang_2021_GDRN}
Wang, G., Manhardt, F., Tombari, F., Ji, X.: {GDR-Net}: Geometry-guided direct regression network for monocular 6d object pose estimation. In: IEEE/CVF Conference on Computer Vision and Pattern Recognition (CVPR). pp. 16611--16621 (June 2021)

\bibitem{wang2019normalized}
Wang, H., Sridhar, S., Huang, J., Valentin, J., Song, S., Guibas, L.J.: Normalized object coordinate space for category-level 6d object pose and size estimation. In: Proceedings of the IEEE/CVF Conference on Computer Vision and Pattern Recognition. pp. 2642--2651 (2019)

\bibitem{wen2021bundletrack}
Wen, B., Bekris, K.: Bundletrack: 6d pose tracking for novel objects without instance or category-level 3d models. In: 2021 IEEE/RSJ International Conference on Intelligent Robots and Systems (IROS). pp. 8067--8074. IEEE (2021)

\bibitem{wen2020se}
Wen, B., Mitash, C., Ren, B., Bekris, K.E.: se (3)-tracknet: Data-driven 6d pose tracking by calibrating image residuals in synthetic domains. In: 2020 IEEE/RSJ International Conference on Intelligent Robots and Systems (IROS). pp. 10367--10373. IEEE (2020)

\bibitem{weng2021captra}
Weng, Y., Wang, H., Zhou, Q., Qin, Y., Duan, Y., Fan, Q., Chen, B., Su, H., Guibas, L.J.: Captra: Category-level pose tracking for rigid and articulated objects from point clouds. In: Proceedings of the IEEE/CVF International Conference on Computer Vision. pp. 13209--13218 (2021)

\bibitem{xiang2017posecnn}
Xiang, Y., Schmidt, T., Narayanan, V., Fox, D.: Posecnn: A convolutional neural network for 6d object pose estimation in cluttered scenes. arXiv preprint arXiv:1711.00199  (2017)

\bibitem{you2022cppf++}
You, Y., He, W., Liu, J., Xiong, H., Wang, W., Lu, C.: Cppf++: Uncertainty-aware sim2real object pose estimation by vote aggregation. arXiv preprint arXiv:2211.13398  (2022)

\bibitem{ze2022category}
Ze, Y., Wang, X.: Category-level 6d object pose estimation in the wild: A semi-supervised learning approach and a new dataset. Advances in Neural Information Processing Systems  \textbf{35},  27469--27483 (2022)

\bibitem{zhang2024omni6dpose}
Zhang, J., Huang, W., Peng, B., Wu, M., Hu, F., Chen, Z., Zhao, B., Dong, H.: Omni6dpose: A benchmark and model for universal 6d object pose estimation and tracking. arXiv preprint arXiv:2406.04316  (2024)

\bibitem{zhang2023genpose}
Zhang, J., Wu, M., Dong, H.: Genpose: Generative category-level object pose estimation via diffusion models. arXiv preprint arXiv:2306.10531  (2023)

\bibitem{zheng2023hs}
Zheng, L., Wang, C., Sun, Y., Dasgupta, E., Chen, H., Leonardis, A., Zhang, W., Chang, H.J.: Hs-pose: Hybrid scope feature extraction for category-level object pose estimation. In: Proceedings of the IEEE/CVF Conference on Computer Vision and Pattern Recognition. pp. 17163--17173 (2023)

\end{thebibliography}
\end{document}